# A Rank based Adaptive Mutation in Genetic Algorithm

Avijit Basak
IBM India Pvt. Ltd.
Kolkata, India

## ABSTRACT

Traditionally Genetic Algorithm has been used for optimization of unimodal and multimodal functions. Earlier researchers worked with constant probabilities of GA control operators like crossover, mutation etc. for tuning the optimization in specific domains. Recent advancements in this field witnessed adaptive approach in probability determination. In Adaptive mutation primarily poor individuals are utilized to explore state space, so mutation probability is usually generated proportionally to the difference between fitness of best chromosome and itself ($f_{MAX}$ - f). However, this approach is susceptible to nature of fitness distribution during optimization.

This paper presents an alternate approach of mutation probability generation using chromosome rank to avoid any susceptibility to fitness distribution. Experiments are done to compare results of simple genetic algorithm (SGA) with constant mutation probability and adaptive approaches within a limited resource constraint for unimodal, multimodal functions and Travelling Salesman Problem (TSP). Measurements are done for average best fitness, number of generations evolved and percentage of global optimum achievements out of several trials. The results demonstrate that the rank-based adaptive mutation approach is superior to fitness-based adaptive approach as well as SGA in a multimodal problem space.

## General Terms

Genetic Algorithm, Adaptive mutation, Fitness distribution, Impact of Skewness in optimization.

## Keywords

Genetic Algorithm, Adaptive mutation.

## 1. INTRODUCTION

In engineering function optimizations of NP-complete category are required in diverse domains. Genetic algorithm based on natural selection is most popular choice to solve these problems. Primarily it works with three different operators i.e. selection, crossover and mutation on a population of randomly created chromosomes, which eventually evolves for multiple generations and converge to a solution. Selection is used to select best chromosomes from the population according to their fitness. Crossover operator is used to recombine the genetic information between two selected chromosomes and generate new offspring with predefined probability. Mutation is used as a secondary operator to alter individual genes of chromosomes with a very low probability. Several approaches were proposed by De Jong (Reference. 2) to improve the basic optimization process like tuning of crossover and mutation probabilities, usage of elitism and increase of population size etc. However, all these works considered constant probabilities for the GA operators. As primary purpose of mutation is to explore the problem space, low probability limits the capability of algorithm noticeably in multimodal domain.

Adaptive mutation approach was designed to overcome this limitation. In adaptive approach mutation probability is varied proportionally to the difference between fitness of best chromosome and itself i.e. ($f_{MAX}$ - f) in a generation. Maximum probability is designed to be constant or vary in an adaptive or controlled way across generations as presented in References 1 and 6. This produced substantial improvement in experimental results for multimodal problem space and achieved global optimum with higher consistency.

However, the adaptive approach based on fitness does not consider the susceptibility of generated probability to nature of fitness distribution. During the process of optimization, the nature of distribution may deviate from symmetry and become skewed. Negatively skewed distribution affects the generated mutation probabilities, eventually slowing the convergence process and sometimes diverting the population from global optimum. This phenomenon becomes more noticeable with smaller population size. Although higher population size can reduce the effect of distribution skewness it also increases cost of optimization. To overcome the problem, in this paper I present an alternative approach of mutation probability generation using chromosome rank.

## 2. PRINCIPLES OF GENETIC ALGORITHM & ADAPTIVE MUTATION

### 2.1 Simple Genetic Algorithm

Genetic algorithm was first proposed by John Holland motivated by Darwin's "Survival of fittest" principle. The algorithm can be viewed as an evolutionary process where a randomly selected population of individuals evolves throughout generations using three operators i.e. selection, crossover and mutation. In each generation highly fit individuals are selected by a suitable selection operator to produce offspring for next generation by crossover. Mutation is used to alter the genetic information in chromosomes to maintain diversity in population. The process continues until the convergence condition is met. The algorithm of SGA is depicted below.

SimpleGeneticAlgorithm () {

initialize population;

evaluate population;

while (convergence condition is not met) {

      select individuals for producing offspring;

      perform crossover to produce offspring;

      perform mutation;

      evaluate population;

}

}

## 2.2 Motivation behind Adaptive Mutation

In simple genetic algorithm there are primarily two reasons for premature convergence, insufficient genetic information in the initial population and loss of genetic information during optimization process. The primary role of mutation is to explore unknown problem space to create new genetic information as well as restore lost ones. In SGA mutation is done with constant probability for all individuals. A higher mutation probability helps in doing better exploration but at the same may lead to information loss from the above average individuals. Ultimately this results in poor quality of convergence.

In adaptive mutation above average individuals in a population are mutated with a very low probability and below average individuals are mutated with high probability. Usually mutation probability is generated proportionally to (fMAX - f) as shown in equation1 below.

**Equation 1:** Fitness-Based Adaptive Mutation Probability

$$p = p_{MAX} * \left(1 - \frac{f}{f_{MAX}}\right)$$

p = Mutation probability of a chromosome

$p_{MAX}$ = Maximum mutation probability

f = Fitness of a chromosome

$f_{MAX}$ = Fitness of the best chromosome in population

## 2.3 Effect of Fitness Distribution on Mutation Probability and Optimization

Stochastic optimization algorithms have dependency on nature of frequency distribution of data it uses. Being stochastic genetic algorithm is not an exception either. In GA nature of fitness distribution keeps on varying over generations during optimization process. Sometimes during optimization very few individuals possess very high fitness which leads to a positively skewed distribution as shown in Figure 1. Contrary to that sometimes few of the individuals possess very low fitness compared to most highly fit individuals and therefore lead to a negatively skewed distribution (Figure 1). During initial stages of optimization, the distribution often becomes positively skewed but as the population approaches towards convergence the distribution often becomes negatively skewed. The effect of positively skewed distribution is usually observed in fitness proportionate selection and that led to the development of other selection approaches like rank selection, tournament selection etc. The effect of negative skewness is more predominant on mutation probability generation using fitness-based adaptive approach described in Equation 1.

According to Equation 1

$$p \to 0 \quad \text{as} \quad \frac{f}{f_{MAX}} \to 1$$

In a negatively skewed distribution majority of individuals possess above average fitness, which leads to generation of very low mutation probabilities. This affects exploration capabilities of those individuals. Closer the optimization approaches the convergence more the negative skewness increases. Gradually the mating chromosomes become nearly identical making crossover almost ineffective. This aggravates the situation even more. Being equipped with ineffective crossover and mutation the optimization process fails to recover the symmetry of distribution and approaches to a local optimum. This phenomenon becomes more dominant for multimodal problem space as higher mutation probability is more desirable to overcome local optima. Lower population size accelerates the problem from very early stage of optimization almost ensuring the convergence failure in a multimodal problem space.

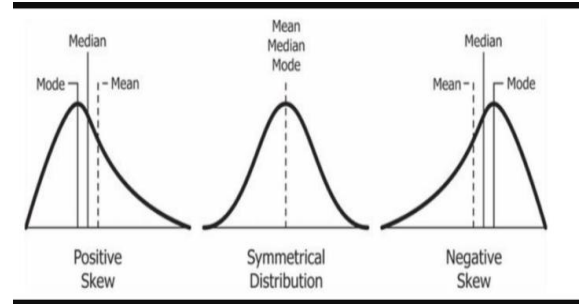

**Figure 1 @source: Wikipedia**

## 3. RANK BASED ADAPTIVE MUTATION

Concerns described in previous section can be overcome by generating mutation probability based on chromosome rank. Rank of a chromosome is decided based on its relative fitness in population. In a population of N individuals' fittest chromosome is assigned a rank N and poorest one is assigned rank 1. All other individuals are assigned a rank between 1 and N based on their fitness value. The normalized rank of chromosome is used to determine the mutation probability as mentioned below in Equation 2.

**Equation 2:** Rank Based Adaptive Mutation Probability

$$p = p_{MAX} * \left(1 - \frac{r - 1}{N - 1}\right)$$

p = Mutation probability of a chromosome

$p_{MAX}$ = Maximum mutation probability

r = Rank of chromosome.

N = Population Size

The equation above ensures that the best chromosome should always have zero mutation probability and the poorest one the highest probability i.e. $p_{MAX}$. The mutation probability of other chromosomes should be distributed linearly between 0 and $p_{MAX}$ based on their rank. However, during optimization if more than one chromosome attains an identical fitness, ranks are assigned randomly to them. As this approach does not use the fitness value of individuals, the mutation probability remains unaffected by the skewness of fitness distribution. A consistent good mutation ensures better exploration capability. In case the optimization process reaches closer to local optimum, the crossover may become less effective but mutation helps to overcome the skewness as well as local optima, henceforth ensuring higher probability of attaining the global optimum.

## 4. EXPERIMENTAL RESULTS AND ANALYSIS

Experimental simulations have been done for both unimodal and multimodal functions and TSP using above mentioned approaches. Results are compared based on few predefined performance measures using a set of common optimization parameters as mentioned below. The purpose of this simulation is to study the effect of distribution skewness in adaptive mutation and demonstrate the benefits of rank-based adaptive mutation. Average mutation probability of .05 is used for SGA and maximum probability of .1 is used for both adaptive approaches. Maximum mutation probability is kept constant throughout generations. In order to study the effect of population size on result, experiments have been performed using population sizes of 10 and 20 respectively. Simulation programs are written in java using apache-commons-math and jfreechart library.

## 4.1 Optimization Strategy

Genetic algorithm uses few parameters to design optimization strategy. The common parameters used for the comparative study are specified below.

SELECTION TYPE: Tournament Selection of size 2

CROSSOVER TYPE: One Point Crossover

CROSSOVER PROBABILITY: 0.8

AVERAGE MUTATION PROBABILITY: 0.05

MAXIMUM MUTATION PROBABILITY: 0.1

## 4.2 Performance Measure

The optimization is performed for below mentioned functions until convergence condition is met. The convergence is assumed to be achieved once best chromosome remains unchanged for 50 consecutive generations. Measurements are collected for average evolution count, average best fitness, maximum evolution count and percentage of global convergence achievement out of 200 trials.

**De Jong's f1:** This is a unimodal quadratic cost function having three independent variables. The function decreases monotonically towards the global optimum coordinate (0,0,0). The problem is encoded as 30-bit binary chromosome. Each independent variable is represented by consecutive 10 binary bits.

$$f_1 = \sum_1^3 x_i^2 \qquad -5.12 \le x_i \le 5.12$$

**f7:** This is a bivariate multimodal cost function having multiple local and one global optimum as mentioned in reference 1. The global optimum is present at coordinate (0,0). The problem is encoded as 24-bit binary chromosome. Each independent variable is represented by consecutive 12 binary bits.

$$f_7 = (x_1^2 + x_2^2)^{.25} * [sin^2(50 * (x_1^2 + x_2^2)^{.1}) + 1.0]$$

$$0 \le x_i \le 40.95$$

**Results for f1:**

**TSP:** Traveling Salesman Problem involves finding out the shortest Hamiltonian cycle in a complete graph of n nodes. Euclidean distance is calculated using coordinates of two cities. In this simulation I have used data of Western Sahara consisting of 29 cities from tsplib. Problem is encoded using random key chromosome of length 29 where order of each random value denotes the seq. of city. Onepoint crossover and Random Key mutation is used for successive population generation.

## 4.3 Results and Analysis

The statistical summary of optimization results and comparative graphical simulations are provided for functions f1 and f7. Table 1 & 2 depicts the comparative statistical data for function f1 using three optimization approaches i.e. SGA, fitness-based adaptive GA and rank-based adaptive GA using different population sizes. Tables 3 & 4 describe the same for function f7. Statistical data for TSP is presented in Tables 5 & 6. The Figure 2, 4 & 6 depicts the variation of lowest cost with generations in a typical trial for functions f1 and f7 and TSP problem respectively. The Figures 3.1 & 2, 5.1 & 2 and 7.1 & 2 present variation of skewness resulting for cost functions f1, f7 and TSP respectively.

From the result it is evident that rank-based mutation probability generation approach produced better result than other approaches. The difference in result is more noticeable for multi-modal function optimization. For both functions it is evident from the diagrams that skewness played the key role behind the quality of optimization. In fitness-based approach the skewness went very low within few generations of evolution which resulted in lower mutation probability for most individuals as described in section 2.C. Although there were some occasional deviations but eventually distribution symmetry could not be recovered. This led to loss of exploration capability and poor optimization quality. However, in rank-based approach mutation probability is not affected by skewness of distribution. Although the skewness went low within few generations it was able to recover and maintain a better average value than other approach. The process also showed quite a good capability of exploring problem space in lesser number of generations. The resulting average cost was quite better than any other approach. Percentage number of global convergence achievement appeared to be quite higher too. However, using larger population size improved the result to some extent for fitness-based approach but failed to meet the quality obtained by rank-based approach. The experimental result proves the effectiveness of the rank-based mutation probability generation approach for multimodal function optimization in adaptive genetic algorithm.

**Table 1**

| Optimization Approach for **f1(population size=10)** | Average Generations evolved | Average Lowest Cost | Global Optimum Achieved | Max No. of Generations evolved | Percentage of global optimum achievement |
|---|---|---|---|---|---|
| SGA | 155.06 | 0.499254 | 48 | 260 | 24.0% |
| Fitness Based AGA | 112.61 | 0.000063 | 186 | 193 | 93.0% |
| Rank Based AGA | 91.00 | 0.00 | 200 | 127 | 100% |

**Table 2**

| Optimization Approach for **f1(population size=20)** | Average Generations evolved | Average Lowest Cost | Global Optimum Achieved | Max No. of Generations evolved | Percentage of global optimum achievement |
|---|---|---|---|---|---|
| SGA | 136.72 | 0.045451 | 119 | 238 | 59.5% |
| Fitness Based AGA | 97.02 | 0.0000005 | 199 | 141 | 99.5% |
| Rank Based AGA | 80.63 | 0.00 | 200 | 99 | 100% |

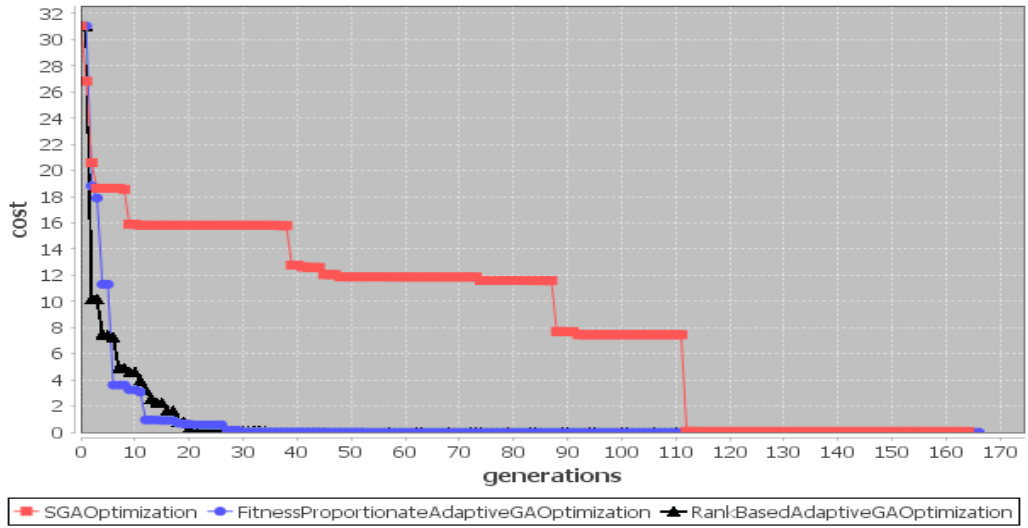

**Figure 1**

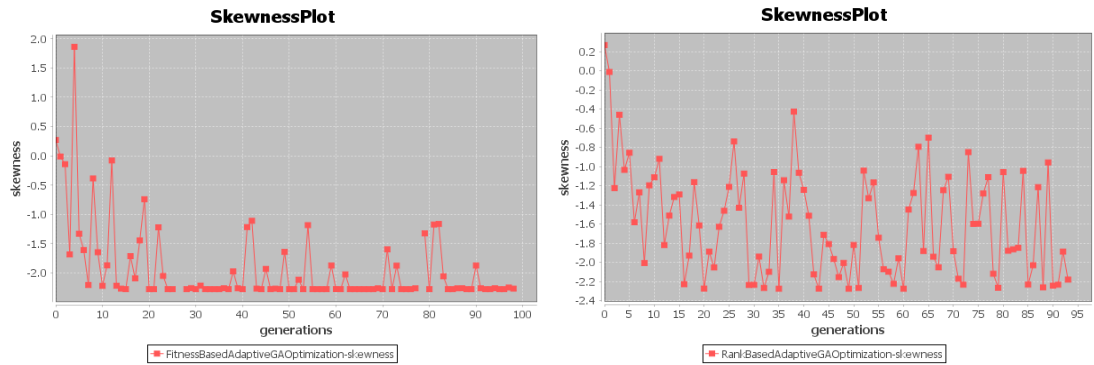

**Figure 2.1**

**Figure 3.2 34.3.2**

**Results for f7:**

**Table 3**

| Optimization Approach for **f7(population size=10)** | Average Generations evolved | Average Lowest Cost | Global Optimum Achieved | Max No. of Generations evolved | Percentage of global optimum achievement |
|---|---|---|---|---|---|
| SGA | 133.97 | 2.2943 | 5 | 324 | 2.5% |
| Fitness Based AGA | 132.51 | 0.5297 | 54 | 299 | 27% |
| Rank Based AGA | 121.99 | 0.0708 | 126 | 245 | 63% |

**Table 4**

| Optimization Approach for *f7(population size=20)* | Average Generations evolved | Average Lowest Cost | Global Optimum Achieved | Max No. of Generations evolved | Percentage of global optimum achievement |
|---|---|---|---|---|---|
| SGA | 120.56 | 1.5012 | 8 | 325 | 4% |
| Fitness Based AGA | 114.82 | 0.3120 | 68 | 251 | 34% |
| Rank Based AGA | 98.57 | 0.0289 | 165 | 164 | 82.5% |

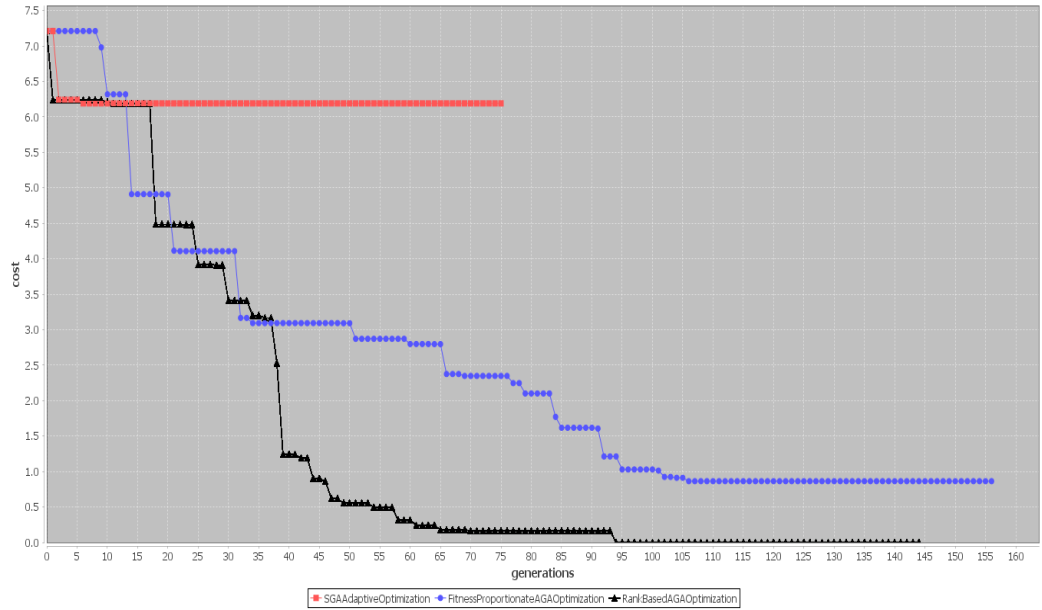

**Figure 4**

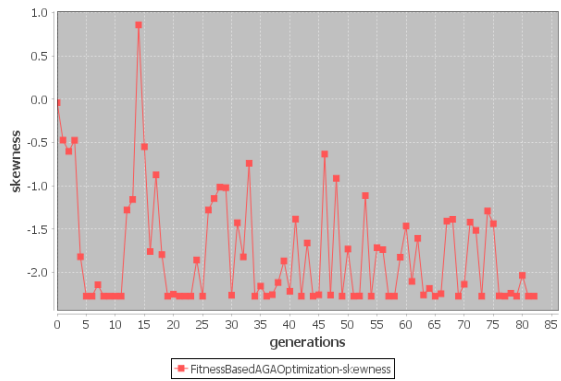

**Figure 5.1**

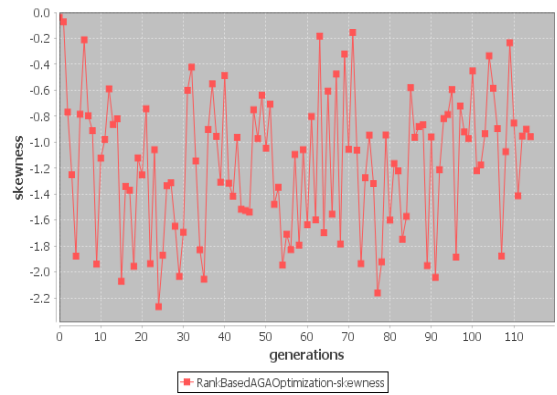

**Figure 5.2**

**Results for TSP:**

**Table 5**

| Optimization Approach for **TSP(population size=250)** | Average Generations evolved | Average Lowest Cost | Global Optimum Achieved | Max No. of Generations evolved | Percentage of global optimum achievement |
|---|---|---|---|---|---|
| SGA | 267.38 | 32067.86 | 0 | 599 | 0% |

| Fitness Based AGA | 237.31 | 31229.51 | 2 | 446 | 1% |
|---|---|---|---|---|---|
| Rank Based AGA | 202.66 | 29780.63 | 15 | 418 | 7.5% |

**Table 6**

| Optimization Approach for **TSP(population size=500)** | Average Generations evolved | Average Lowest Cost | Global Optimum Achieved | Max No. of Generations evolved | Percentage of global optimum achievement |
|---|---|---|---|---|---|
| SGA | 234.19 | 30867.00 | 3 | 510 | 1.5% |
| Fitness Based AGA | 214.49 | 29873.39 | 10 | 515 | 5% |
| Rank Based AGA | 175.82 | 29290.91 | 21 | 368 | 10.5% |

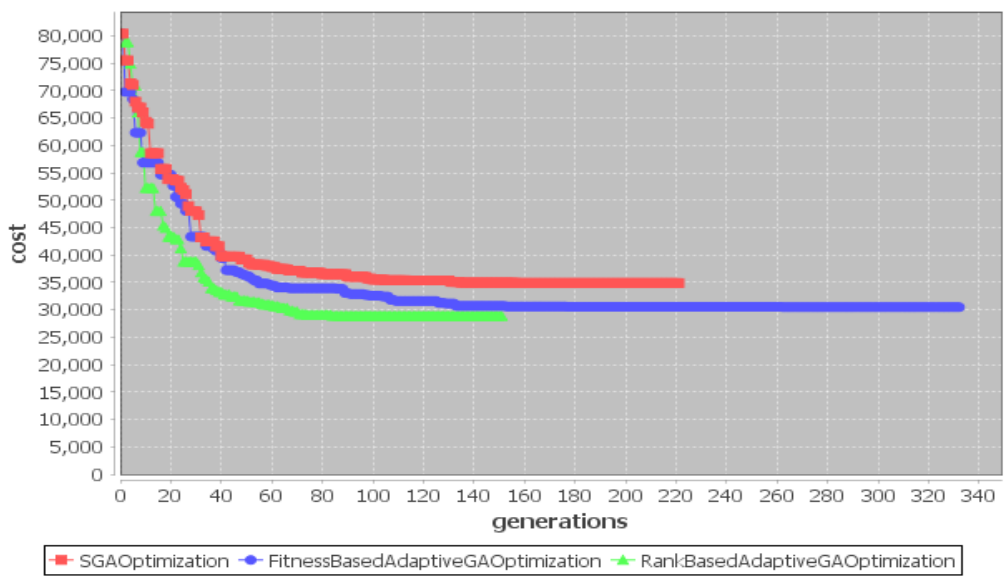

**Figure 5.2**

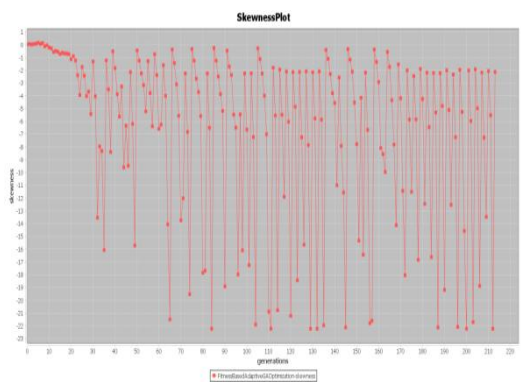

**Figure 7.1**

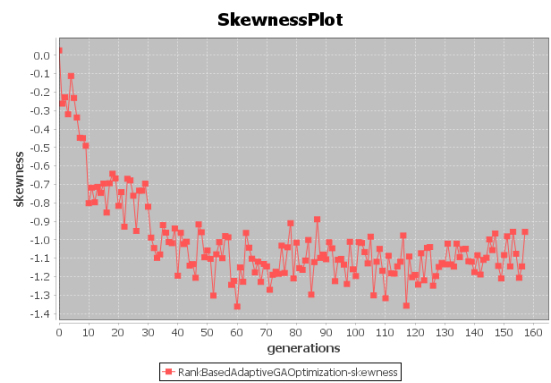

**Figure 7.2**

# 5. CONCLUSION

Recent researches in GAs have incorporated adaptive approach for different operators. This paper demonstrates an approach of adaptive mutation based on chromosome rank and compares the same with other similar approaches. The effect of fitness distribution on the generated mutation probability is studied here for both adaptive approaches. After performing rigorous experimental studies, the rank-based approach proved to be more effective for mutation probability generation in GA. The better quality of optimization is achieved with lower resource cost. This approach should be particularly useful for optimization in resource intensive applications. Similar approach can be used for adaptive crossover also. Adoption of this approach along with parallel genetic algorithm will improve convergence quality to a greater extent and further research can be done in this area.

# 6. REFERENCES

[1] Srinivas, M. and Patnaik, L. M., Adaptive Probabilities of Crossover and Mutation in Genetic Algorithms, IEEE Transactions on Systems, Man and Cybernetics, VOL. 24, NO. 4, APRIL 1994.

[2] DeJong, K. A., An analysis of the behaviour of a class of genetic adaptive systems Ph.D. dissertation, University of Michigan (1975)

[3] Goldberg, D. E., Genetic Algorithms in Search, Optimization and Machine Learning, Addition-Wesley Publishing Co., Inc. Boston, MA, USA. (1989)

[4] Mille, Brad L., Goldberg, D. E., Genetic Algorithms, Tournament Selection, and the Effects of Noise, Complex Systems 9 (1995) 193- 212

[5] Goldberg, D.E., Deb, K., A Comparative Analysis of Selection Schemes Used in Genetic Algorithms, University of Illinois at Urbana-Champaign, Urbana, United States

[6] Kühn, M., Severin, T., Salzwedel H., Variable Mutation Rate at Genetic Algorithms: Introduction of Chromosome Fitness in Connection with Multi-Chromosome Representation, International Journal of Computer Applications (0975 – 8887), Volume 72–No.17, June 2013

[7] Khemani, Deepak, A First Course in Artificial Intelligence, McGraw Hill Education (India) Private Limited, Chennai, India

[8] Korejo++, I.A., Khuhro, Z.U.A., Jokhio, F. A., Channa*, N., Nizamani, H. A., An Adaptive Crossover Operator for Genetic Algorithms to Solve the Optimization Problems, Sindh University Research Journal (Science Series) Vol.45 (2) 333-340 (2013) Ding, W. and Marchionini, G. 1997 A Study on Video Browsing Strategies. Technical Report. University of Maryland at College Park.